\pdfoutput=1

\documentclass[11pt]{article}

\usepackage[]{latex/acl}

\usepackage{times}
\usepackage{latexsym}

\usepackage[T1]{fontenc}

\usepackage[utf8]{inputenc}

\usepackage{microtype}

\usepackage{inconsolata}

\usepackage{booktabs} 
\usepackage{graphicx}
\usepackage{amsmath,amsfonts,amssymb}

\usepackage{xcolor}         
\usepackage{amsmath}
\usepackage{soul}

\usepackage{subcaption}
\usepackage{cinzel} 
\usepackage{multirow}
\usepackage{multicol}
\usepackage{tabularx}

\usepackage{comment} 
\usepackage{wrapfig} 
\usepackage{bbm}
\usepackage{pifont}
\usepackage{makecell,rotating}
    \setcellgapes{5pt}
\usepackage{calligra}

\usepackage{aurical}

\usepackage{listings}
\usepackage{color}
\usepackage[T1]{fontenc}
\usepackage{numprint}

\npthousandsep{\,}
\npdecimalsign{.}
\npproductsign{\cdot}
\npunitseparator{}

\definecolor{dkgreen}{rgb}{0,0.6,0}
\definecolor{gray}{rgb}{0.5,0.5,0.5}
\definecolor{mauve}{rgb}{0.58,0,0.82}

\usepackage{titlesec}
\titlespacing*{\subsection}
{0pt}{0ex}{0ex}
\titlespacing*{\subsubsection}
{0pt}{0ex}{0ex}

\usepackage{mathtools}
\usepackage{titlesec}

\title{Enhancing Hallucination Detection through Perturbation-Based Synthetic Data Generation in System Responses}

\author{Dongxu Zhang, Varun Gangal, Barrett Martin Lattimer, Yi Yang\\
ASAPP, Inc.\\
  \texttt{\{dzhang,vgangal,blattimer,yyang\}@asapp.com}}

\begin{document}
\maketitle
\begin{abstract}

Detecting hallucinations in large language model (LLM) outputs is pivotal, yet traditional fine-tuning for this classification task is impeded by the expensive and quickly outdated annotation process, especially across numerous vertical domains and in the face of rapid LLM advancements. In this study, we introduce an approach that automatically generates both faithful and hallucinated outputs by rewriting system responses. Experimental findings demonstrate that a T5-base model, fine-tuned on our generated dataset, surpasses state-of-the-art zero-shot detectors and existing synthetic generation methods in both accuracy and latency, indicating efficacy of our approach.

\end{abstract}

\section{Introduction}


Large Language Models (LLMs) tend to produce hallucinations, wherein the generated text either contradicts the given source knowledge (intrinsic hallucination) or cannot be verified against it (extrinsic hallucination)~\cite{maynez2020faithfulness,rawte2023survey}. Despite the burgeoning enthusiasm for deploying Generative AI and LLMs in real-world applications, the issue of hallucinations poses significant concerns for downstream users. Consequently, the detection of hallucinations is paramount in enhancing the safety of LLM applications and in fostering trust among users of these technologies.


An effective hallucination detection system should be accurate, fast, and affordable. Cost-effectiveness is crucial because every check for hallucinations adds extra cost to the use of large language models (LLMs), which may already be substantially high. Moreover, the system must possess the flexibility to adapt to the rapidly evolving landscape of LLMs. As shown in Table~\ref{tab:outdate_performance}, newer iterations of LLMs generally exhibit enhanced capabilities in mitigating hallucinations, thereby escalating the complexity of the detection challenge. Unfortunately, many current methodologies are either i) costly in terms of compute \cite{liu-etal-2023-g,manakul2023selfcheckgpt}
or ii) depend on out-of-domain/external resources such as QA \cite{honovich-etal-2021-q2,fabbri2022qafacteval} or NLI annotation ~\cite{laban-etal-2022-summac,honovich2022true}, potentially compromising performance.


\begin{table}[htb]
\caption{Performance evaluation of a GPT-3.5-based zero-shot hallucination detector across different generations of LLMs (see Appendix \S E for prompt). This table illustrates a notable decline in detection efficacy when transitioning from older to more recent LLM iterations.} 
\label{tab:outdate_performance}
\centering
\small
\begin{tabular}{l|r|r}
\toprule 
Hallucination data & LLMs used in the data & F1 \\
\midrule
MNBM ('20)  & GPT, Bert, Rnn, ConvNet & $0.780$  \\
FRANK ('21) &PointerNet, bertS2S, Bart  & $0.694$  \\
Seahorse (early '23)  & T5, MT5, PALM & $0.576$  \\
ScreenEval (late '23) & GPT-4, longformer  & $0.130$  \\
\bottomrule
\end{tabular}
\end{table}

\begin{figure}[h]
    \centering
    \includegraphics[width=7cm]{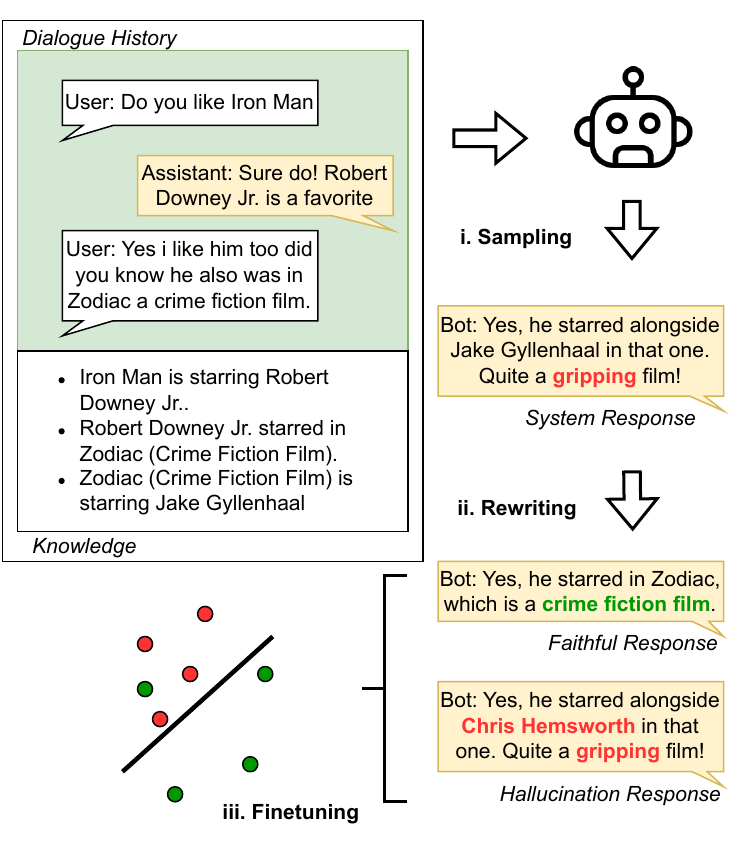}
    \caption{Overview of our automatic hallucination generation pipeline. Red and green highlights hallucinated and faithful claims.} 
    \label{fig:diagram}
\end{figure}

In this study, we introduce a simple yet effective approach for automatically generating synthetic annotations to train hallucination detectors. Figure~\ref{fig:diagram} shows an overview of our approach. The core of our method involves prompting a rewriting LLM to transform a given system response from the target LLM into both faithful and hallucinated versions, respectively. This technique distinguishes itself from existing methods~\cite{gupta2021dialfact,das2022diving,li-etal-2023-halueval,dziri-etal-2022-evaluating} in three significant ways. First, unlike traditional methods that rely on human-annotated examples of faithfulness, our strategy is entirely automated, eliminating  need for manual annotation. Second, by directly altering responses from the target LLM, our trained detector aligns more closely with the response distribution of the target LLM, facilitating seamless adaptation to new LLMs. Lastly, while previous approaches require predefined information about the types of hallucinations for their generation process, our method operates without such assumptions. This allows for the creation of a broader spectrum of hallucination types, enhancing the coverage and diversity of generated hallucinations.

Our experimental evaluations span two hallucination detection datasets, OpenDialKG~\cite{moon-etal-2019-opendialkg} and BEGIN~\cite{dziri2022evaluating}, where a T5-base model, fine-tuned with our novel data generation approach, significantly surpasses GPT-4 based methods in performance while achieving a tenfold increase in speed. Further analysis of the generated hallucinations uncovers previously unreported patterns, such as "adding attributes", expanding the discourse on hallucination beyond existing literature. Our code and data will be available at \url{https://github.com/asappresearch/halugen}.

\section {Methodology}
\label{sec:methodology}

In this section, we detail our methodology for generating synthetic hallucinations that closely mimic those encountered in real-world applications of Large Language Models (LLMs). Prior approaches to hallucination generation have primarily relied on rewriting human-authored texts~\cite{das2022diving,li-etal-2023-halueval} or introducing perturbations to the knowledge source~\cite{gupta2021dialfact,dziri-etal-2022-evaluating,zhang2023alleviating}. However, these methods often yield outputs that diverge significantly from those produced by LLM systems, leading to a substantial discrepancy between the synthetic hallucinations and the genuine hallucinations observed in practice. To address this gap, our approach involves prompting a rewriting LLM to perturb the responses of the LLM system itself, rather than those written by humans. This strategy draws inspiration from the ``Minor perturbation'' technique described by~\newcite{lucas-etal-2023-fighting}, adapted to our context to ensure the synthetic hallucinations closely align with the expected data in real-world deployments.

To effectively train a hallucination detector, it is imperative to have access to both hallucinated and faithful responses. Unlike previous studies, where human-curated outputs served as the benchmark for faithful system outputs~\cite{das2022diving,li-etal-2023-halueval,dziri-etal-2022-evaluating}, the responses obtained directly from the target LLM system may contain a considerable proportion of non-faithful responses. To overcome this challenge, we employ the rewriting LLM to adjust the system's responses in a manner that promotes the generation of faithful outputs. The specific prompts utilized for inducing both hallucination and faithfulness are presented in Appendix \S\ref{sec:appendix_generation_prompt}. It is important to note that our process for generating hallucinations did not involve biasing the system with predefined categories of hallucination within the prompt, ensuring a more authentic and unbiased generation process.\footnote{These prompts have been designed with versatility in mind, allowing for straightforward adaptation to other NLP tasks such as question answering and summarization. However, our current investigation is focused exclusively on knowledge-grounded dialogues.} For the rewriting LLM, we selected GPT-4~\footnote{We use gpt-4-1106-preview for our experiments.} due to its robust capabilities in text rewriting \cite{madaan2024self}. Leveraging a powerful rewriting LLM like GPT-4 enables the exploration of a wider array of hallucination categories, thereby enhancing the coverage of hallucinations that are likely to be encountered in real-world scenarios.

\section{Experiments}
\label{sec:results_and_analysis}
\subsection{Datasets}

\paragraph{OpenDialKG} is a dialogue dataset that was adopted by HaluEval~\cite{li-etal-2023-halueval}, a recent benchmark for hallucination detection. OpenDialKG features human-generated dialogues exclusively with supporting knowledge sources from Freebase \cite{bollacker2008freebase}. In order to leverage the dataset for hallucination detection, we simulate a chatbot system by employing GPT-4 to generate responses grounded in both the provided knowledge and the preceding dialogue context. The specifics of the prompt template utilized for this simulation are detailed in Appendix \ref{sec:appendix_simulation_prompt}. To create a evaluation set on the generated responses, we employ Amazon Mechanical Turk annotators to evaluate whether the responses from the simulated chatbot system were fully supported by the dialogue history and the provided knowledge (for detailed annotation guidelines and interface, refer to Appendix D). Our collection (OpenDialKG-Eval) comprises 402 annotated responses. We designated responses with high-confidence labels as our test set and utilized the remainder for development purposes, resulting in 312 test responses and 90 for development. More details of OpenDialKG-Eval can be found in Appendix ~\ref{appendix:data_stats}. In addition, we simulate another 2000 responses from OpenDialKG for synthetic generation purpose.

\paragraph{BEGIN} is a knowledge-grounded dialog dataset featuring 12k responses from four dialogue systems distributed over 3 document-scale knowledge domains -- Wizard of Wikipedia \cite{dinan2018wizard}, TopicalChat \cite{gopalakrishnan2023topical} and DoG~\cite{zhou2018dataset} --- all with mean knowledge snippets longer than OpenDialKG. 
In addition, there are three response categories in BEGIN: Fully attributable, Not fully attributable, Generic. Generic category refers to response that are vague and do not provide any new information. Therefore, in addition to faithful and hallucination generation, we also ask LLM to generate responses under "Generic" category. The detailed prompt can be found in Appendix~\ref{sec:appendix_generation_prompt} Table~\ref{tab:prompt_generic_generation}.
Since BEGIN only released the Dev and Test split, we adopt 1,228 system responses from Dev for both synthetic generation and development while reporting results on Test split.

\subsection{Baselines}
\paragraph{Zero-shot Detection}
We compare with SelfCheckGPT~\cite{manakul-etal-2023-selfcheckgpt}, a consistency-based approach which samples system responses multiple times in temperature 1.0 and then leverage scores from NLI or QA to measure whether the target response is consistent with these samples. 

Another baseline is G-Eval \cite{liu-etal-2023-g}, which prompts GPT-4 with an annotation-rubric style prompt describing target variable and furthermore draws multiple samples at a higher temperature; emulating diverse multi-annotation by humans.
Since both G-Eval and SelfCheckGPT can only output scores between 0 and 1 and BEGIN data has three output categories, we compare GPT-4 (Internal), our self-devised zero-shot detector, which prompts GPT-4 with an intuitive prompt to enable three-way outputs(Appendix\S F) and does greedy decodes to generate a binary/ternary answer. 

The last zero-shot baseline we compare with is SCALE \cite{lattimer2023fast}, NLI-based approach which first decomposes the supporting context into chunks, calculate NLI scores on the chunk level using FlanT5~\cite{chung2022scaling}, then use the maximum score as the final prediction of factual consistency.

\paragraph{Detection with End-to-end Finetuning} 

We use T5-base, an encoder-decoder LM with 223M parameters, as the base model of the detector and fine-tune it on multiple synthetic datasets.\footnote{For more experimental details, please refer to Appendix~\ref{appendix:finetune}.}
We make our best efforts to conduct apple-to-apple comparison among different synthetic data. 
On OpenDialKG-Eval benchmark, we compare with FADE~\cite{das2022diving} and HaluEval~\cite{li-etal-2023-halueval}, where we adopted their existing synthetic hallucinations as negative and human written responses from OpenDialKG as positive data for training.
On BEGIN dataset, we compare with AugWOW~\cite{gupta2021dialfact} and BEGIN-Adv.~\cite{dziri-etal-2022-evaluating}, both are synthetic generation baselines and their performances on BEGIN has been reported by ~\cite{dziri-etal-2022-evaluating}. For more details of these synthetic data generation baselines, please refer to Section~\ref{sec:background}

\subsection{Results}






\begin{table}[h]
\caption{Macro-F1 and latency of hallucination detection methods over OpenDialKG-Eval. } 
\label{tab:perf_opendialkg}
\centering
\small
\begin{tabular}{l|r|r}
\toprule

& F1 & Latency \\
\midrule
\multicolumn{3}{c}{Zero-shot Detection} \\
\midrule
\scriptsize{SelfCheckGPT (QA)} \cite{manakul-etal-2023-selfcheckgpt} & 0.536 & 60.59 sec \\
\scriptsize{SelfCheckGPT (NLI)} \cite{manakul-etal-2023-selfcheckgpt} & 0.579 & 0.93 sec \\
G-Eval \scriptsize{\cite{liu-etal-2023-g}} & 0.608  & 2.79 sec \\
SCALE$_{XL}$ \scriptsize{\cite{lattimer2023fast}} & 0.687  & 0.22 sec  \\

\midrule
\multicolumn{3}{c}{{  }T5-base Finetuned over Synthetic Data} \\
\midrule

FADE \scriptsize{\cite{das2022diving}}  & 0.625 &  0.20 sec\\ 
HaluEval \scriptsize{\cite{li-etal-2023-halueval}}  & 0.702 &  0.20 sec\\ 
Our approach  & 0.762 & 0.20 sec  \\

\bottomrule
\end{tabular}
\end{table}

Table~\ref{tab:perf_opendialkg} shows the performance of hallucination detection and latency per response on OpenDialKG-Eval. Latencies are profiled over AWS g5.xlarge instances with no batching sae for G-Eval which requires OpenAI API access.
From the results, our approach not only out-performs T5 detectors finetuned over previous hallucination generation baselines, but more interestingly, it out-performs state-of-the-art zero-shot detection methods. Besides performance, finetuned models achieve significantly lower latency than all zero-shot baselines. We also show the results on BEGIN data. The results can be found in Table~\ref{tab:perf_begin}, where similar observation can be found. 

\begin{table}[h]
\caption{Macro-F1 and latency of hallucination detection over BEGIN test split with three-class classification. 
} 
\label{tab:perf_begin}
\centering
\small
\begin{tabular}{l|r|r}
\toprule
& F1 & Latency \\
\midrule
\multicolumn{2}{c}{Zero-shot Detection} \\
\midrule
GPT-4 (Internal) & 0.323 & 1.13 sec \\
\midrule
\multicolumn{2}{c}{T5-base Finetuned over Synthetic Data} \\
\midrule


AugWow \scriptsize{\cite{gupta2021dialfact}}  &  0.378 &  0.20 sec\\
BEGIN-Adv. \scriptsize{\cite{dziri-etal-2022-evaluating}} & 0.459 & 0.20 sec\\
Our approach  & 0.473 & 0.20 sec\\

\bottomrule
\end{tabular}
\end{table}

Lastly, average cost per synthetic response generation is 0.008 USD on OpenDialKG and 0.006 USD on BEGIN, using \emph{gpt-4-1106-preview}. In comparison, average cost of human annotation per example for OpenDialKG-Eval is 0.20 USD.

\subsection{Ablation Study}

To analyze the significance of both hallucination and faithful response generation, we conduct an ablation study to replace one of the generation using system response. Results are shown in Table~\ref{tab:ablation_study}. Results show that both categories of synthetic data are necessary to effectively fine-tune the detector.

\begin{table}[h]
\caption{Results of ablation study. ``pos-F1'' and ``neg-F1'' represents F1 performance over faithful and Hallucination labels separately.
} 
\label{tab:ablation_study}
\centering
\small
\begin{tabular}{l|r|r|r}
\toprule
Approach & pos-F1 & neg-F1 & F1 \\
\midrule
Our approach & 0.812 & 0.713 & 0.762 \\
w/o faithful generation &  0.747 & 0.618 & 0.683 \\
w/o hallucination generation & 0.517 & 0.502 & 0.509 \\
\bottomrule
\end{tabular}
\end{table}

\section{Hallucination Pattern Analysis} 
\label{sec:Analysis}
\subsection{Hallucination Pattern Analysis} 
\label{subsec:HalluPat}
Previous work usually predefined hallucination patterns such as replacing or swapping entities~\cite{das-etal-2022-diving,li-etal-2023-halueval}. We randomly sample 144 hallucinations generated by our method over OpenDialKG dataset, and manually annotate these into a taxonomy of 6 distinct pattern-driven categories characterizing the pattern surfaced in the hallucination, further described in Appendix \S C.

\begin{table}[htb]
\caption{Hallucination patterns appeared in OpenDialKG-Eval and our synthetic generated data for finetuning.} 
\label{tab:hallucination_patterns}
\centering
\tiny
\begin{tabular}{l|r|r|r|r}
\toprule
Pattern name  & System  & HaluEval & FADE  & Ours \#\\
\midrule
Adding attribute to an entity & 0.540   & \textbf{0.435} & 0.156   & 0.530 \\
Adding or updating relation & 0.070    & 0.150 & 0.099 &  0.220\\
Addding new entities & 0.050   & \textbf{0.370} & \textbf{0.675} & 0.160 \\
Overclaim knowledge/affordance & 0.027  & 0.011  & 0.010 & 0.025 \\
Inference error beyond above & 0.004 & 0.011  & 0.018 & 0.010 \\
None of the above & 0.310 & 0.016  & 0.042 & 0.050  \\ \midrule 
$\text{KL}(\bullet,\text{System})$ & - & 0.671 & 1.527 & 0.340 \\
\bottomrule
\end{tabular}
\vspace{-1mm}
\end{table}

\begin{figure}[t]
    \centering
    \includegraphics[width=0.4\textwidth]{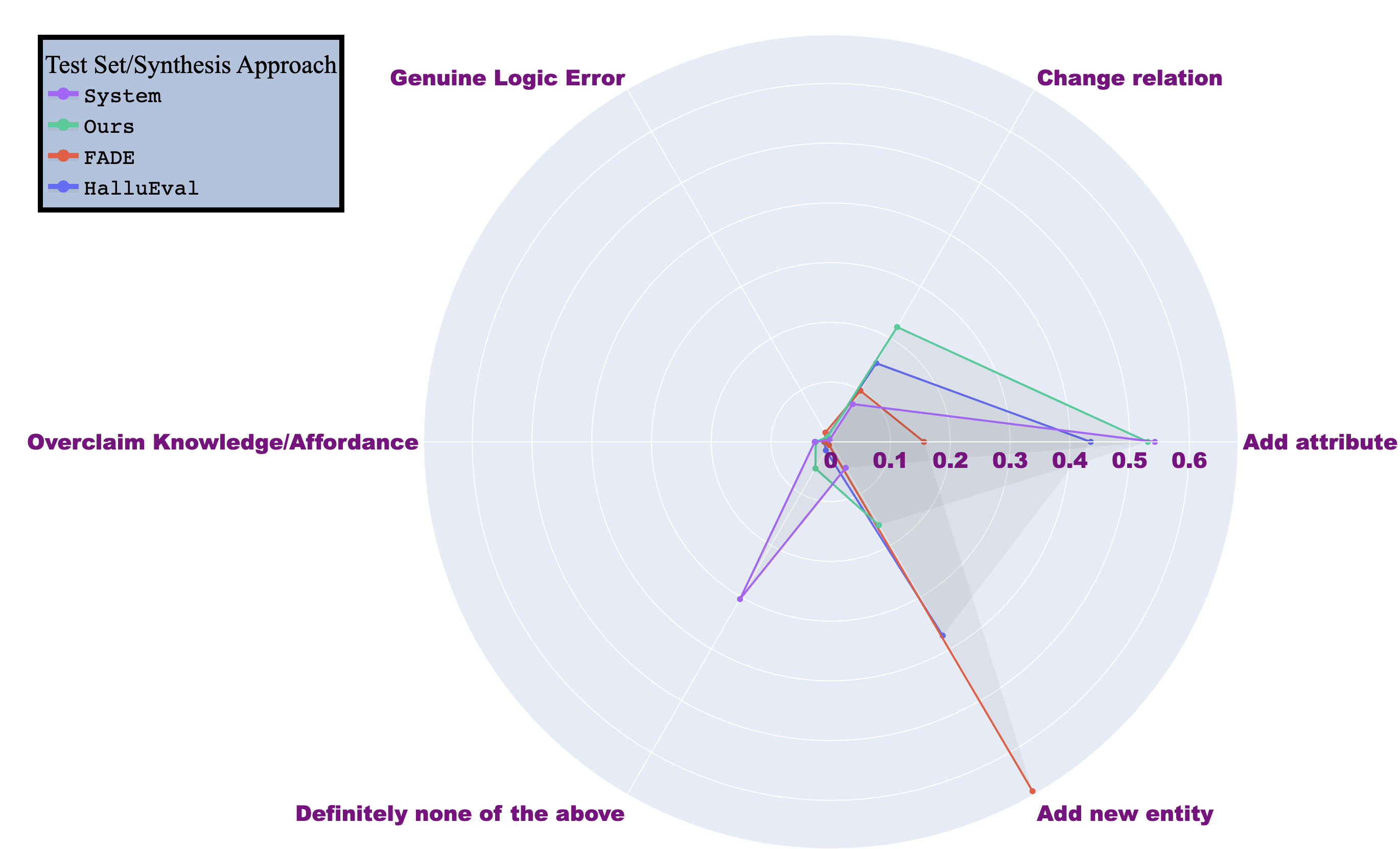}
    \caption{\footnotesize Spiderplot spider-web traces visualizing how the synthesized hallucinations from our approach (in \textit{green}) + two baselines (\textsc{HaluEval,Fade}, in \textit{red} and \textit{blue}) as well as the system response distribution (\textsc{System},in \textit{purple}) distribute over the 6 qualitative categories as laid out in \S\ref{subsec:HalluPat}. Both \textsc{HaluEval} (blue) and \textsc{FADE} (red) show a marked skew towards "Add new entity", while \textsc{Ours} (green) shows a closer alignment with the \textsc{System} (purple).} \label{fig:spiderPlotPatternDistrib}
\end{figure}

Pattern distributions are both listed in Table~
\ref{tab:hallucination_patterns} and Figure~\ref{fig:spiderPlotPatternDistrib}. From the pattern distribution, it is interesting to see that our method has fewer hallucinations from entity replacing/swapping, the most dominant hallucination type is adding unverifiable attributes to an entity. This indicates that our methods generate responses which conform tighter to the real hallucination distribution in contrast to prior approaches. The KL Divergence between the categorical pattern distribution of our method and the system response based  distribution is 0.3395, compared to the much greater 0.6706 (and 1.52) between the distribution of HaluEval (and FADE) vs the latter. 

\subsection{Quality Analysis of Synthetic Data Generation} 
\label{subsec:QualityAnalysis}

To more closely evaluate the effectiveness of rewriting, we did human annotation over 100 randomly sampled system responses along with our synthetically generated responses based on these system responses. Table ~\ref{tab:quality_analysis} shows the portion of faithful data within each type of responses:

\begin{table}[h]
\caption{Human analysis of faithfulness of our generated synthetic responses in comparison to system outputs.
} 

\label{tab:quality_analysis}
\centering
\small
\begin{tabular}{l|r}
\toprule
& Faithfulness \\
\midrule
System output & 41\% \\
\midrule
Faithful generation & 51\% \\
Hallucination generation & 5\% \\
\bottomrule
\end{tabular}
\end{table}

Upon reviewing the annotations, our method demonstrates a significant reduction in the generation of unfaithful responses (hallucinations) compared to system outputs. Our approach also yields a higher number of faithful responses compared to the baseline system outputs, aligning with our objectives. Specifically, out of 59 instances of hallucinations identified in system outputs, our faithful generator converts 11 into faithful responses, achieving a conversion rate of approximately 19\%. Conversely, among the 41 faithful responses generated by the system, our method inadvertently transformed only one into a hallucination.

\section {Related Work}
\label{sec:background}

\label{subsec:synDataGenHalluDetect}

Research on generating synthetic annotations for hallucination detection has explored various strategies. Some approaches, like FADE~\cite{das2022diving} and HaluEval~\cite{li-etal-2023-halueval}, manipulate human-written texts by altering entities or applying predefined hallucination criteria, respectively. These methods 
only focus on introducing hallucinations and ignore faithfulness augmentation, assuming human-generated content to be inherently accurate, which maybe untrue. 
Other studies focus on modifying the knowledge source before response generation. AugWow~\cite{gupta2021dialfact} introduces hallucinations by using irrelevant or no evidence, while BEGIN-Adv~\cite{dziri-etal-2022-evaluating} alters subjects, objects, named entities, or verbs in the source material, prompting a GPT2-based system~\cite{radford2019language} for response regeneration. These techniques, however, might lead to predictable hallucination patterns due to their reliance on predefined rules.

More recently, 
ICD \cite{zhang2023alleviating} mitigates LLM hallucinations by finetuning a model on non-factual samples, aiming to down-weight factually weak predictions. Despite its novelty, the reliance on entity perturbation for generating non-factual samples could limit the coverage of detected hallucinations. 
HaluEval-Wild~\cite{HaluEval-Wild} aims to evaluate LLM hallucination in human-LLM interactions. Their approach first collects challenging user queries which can lead to hallucinated LLM responses. Faithful reference responses are generated using GPT4 with retrieval augmentation. 
While the generated data is challenging, it is not obvious how to adapt the approach for customized tasks.
Interestingly, ~\newcite{li2023synthetic} observed that the effectiveness of the LLM-generated synthetic data in supporting model training is negatively correlated with the subjectivity of the target task.

\section{Conclusions}
\label{sec:conclusion_future_work}
In this work, we aim to address the prevalent challenge of training data for hallucination detection being either unavailable or expensive to curate. We hypothesize that this can be addressed via a framework that automatically synthesizes both hallucinated and faithful responses using a prompt-based method. Our experimental results on two datasets verify effectiveness of our approach and show it compares favourably against several baselines, including those using prompt-based synthesis.


\section{Limitations}
In this work, the quality of the synthetically generated data is partially determined by the capability of prompted LLM. However, this issue is not severe since our goal is to facilitate the fine-tuning process of the hallucination detection model rather than using the data for evaluation. 

In addition, note that hallucinations generated in this work, may still be different from those exist unintentionally in system outputs. One promising future work is to explore unintentional hallucination/faithful output generation, such as evaluating output faithfulness via sampling, or perturbing the prompt so that it becomes more or less likely to induce hallucinations.

Lastly, since we are encouraging the LLM to generate hallucinations, there is a risk of introducing misinformation into the real world data, which is also a common issue for large language model generation in general. We encourage people to follow policies and strategies with regarding to data sourcing, fact checking, etc. in order to mitigate such issue.



\bibliography{anthology,custom}

\clearpage
\appendix

\section{Prompt Template for Synthetic Response Generation}
\label{sec:appendix_generation_prompt}

Table~\ref{tab:prompt_hallucination_generation} and Table~\ref{tab:prompt_faithful_generation} include the prompt templates to generate hallucinated responses and faithful responses.

For BEGIN dataset, we also create a prompt to generate "Generic" responses, as shown in Table~\ref{tab:prompt_generic_generation}

\begin{table*}[h]
\caption{Prompt template to generate hallucinated responses. 
} 
\label{tab:prompt_hallucination_generation}
\centering
\small
\begin{tabular}{l}
\toprule

Take a deep breath and work on this problem step by step. \\
I want you act as a chatbot in a conversation with human. Your job is to edit a detail in the True Response and generate\\
a Hallucinated Response that is inconsistent with the Dialogue History and Knowledge.\\
- Valid edit actions include removing, replacing or adding a short piece of information to the True \\
Response.\\
- If the True Response is faithful, please edit it to generate a Hallucinated Response.\\
- If the True Response has already contained hallucination, please edit it to generate an adversarial Hallucinated Response \\
that are more difficult to be detected.\\
- The generated Hallucinated Response should be ambiguous or complex or non-trivially implicit to be detected by a \\ 
human who has access to all the Knowledge and Dialogue History. \\
- The generated Hallucinated Response should contain similar number of words as the True Response. Do not make it \\
lengthy.\\
\\
\#Knowledge\#: \{Instructional prompt for target system\}\\
\#Dialogue History\#: \{dialogue history\}\\
\#True Response\#: \{system output\}\\
Now, please generate your hallucinated response:\\
\#Hallucinated Response\#:\\

\bottomrule
\end{tabular}
\end{table*}

\begin{table*}[h]
\caption{Prompt template to generate faithful responses. 
} 
\label{tab:prompt_faithful_generation}
\centering
\small
\begin{tabular}{l}
\toprule
Take a deep breath and work on this problem step by step. \\
I want you act as a chatbot in a conversation with human. \\
Given a Response that contains hallucination, your job is to edit the Response lightly and generate a faithful Response \\
that is fully supported by with the Dialogue History and Knowledge.\\
- Valid edit actions include removing or replacing a short piece of information to the Response.\\
- Every token of the generated Response should be strictly verifiable by the Knowledge and Dialogue History. Even\\
commonsense information needs to be verifiable.\\
- Please keep the similar writing style as the Response. Do not make your response lengthy.\\
\\
\#Knowledge\#: \{Instructional prompt for target system\} \\
\#Dialogue History\#: \{dialogue history\}\\
\#Response\#:\{system output\}\\
\\
Now, please generate your faithful response:\\
\#Faithful Response\#:\\

\bottomrule
\end{tabular}
\end{table*}

\begin{table*}[h]
\caption{Prompt template to generate 'Generic' responses for BEGIN dataset. 
} 
\label{tab:prompt_generic_generation}
\centering
\small
\begin{tabular}{l}
\toprule
Take a deep breath and work on this problem step by step. \\
I want you act as a chatbot in a conversation with human. \\
Given a Response, your job is to rewrite it such that it is ostensibly about the same topic as the Response but becomes \\
vague and does not contain any factual statement. \\
Examples of rewritten Response includes but not limited to back-channeling, expressing uncertainty, or diverting the \\
conversation from ambiguous or controversial topics.
Do not make your response lengthy.\\
\\
\#Knowledge\#: \{Instructional prompt for target system\}\\
\#Dialogue History\#: \{dialogue history\}\\
\#Response\#: \{system output\}\\
Now, please generate your faithful response:\\
\#Rewritten Response\#:\\

\bottomrule
\end{tabular}
\end{table*}

\section{Prompt Template for Simulating Chatbot on OpenDialKG}
\label{sec:appendix_simulation_prompt}

Table~\ref{tab:prompt_simulation} contains the prompt template that we use to prompt GPT-4 for system responses on OpenDialKG.

\begin{table*}[h]
\caption{Prompt template to simulate the chatbot system for OpenDialKG. 
} 
\label{tab:prompt_simulation}
\centering
\small
\begin{tabular}{l}
\toprule
Take a deep breath and work on this problem step by step. \\
Given a Dialogue History and Knowledge, your job is to follow instructions in the Knowledge and generate a faithful \\
Response based on the Knowledge and Dialogue History.\\
\#Knowledge\#:\\ 
You are a chatbot. Your goal is to continue the conversation by responding to user's last utterance.\\
\\
You have the following knowledge that can be used to generate your response:\\
\{KG knowledge\}\\
\#Dialogue History\#: \\
\{dialogue history\}\\
Now, please generate your response:\\
\#Response\#: \\

\bottomrule
\end{tabular}
\end{table*}

\section{Rubric/Typology for Qualitative Annotation}
For the qualitative annotation in Table 4 of the main body, we use the rough definitions/guidelines below. We formulate these types based on prior work on hallucination and hallucination typology such as FRANK.

\begin{itemize}
\item Type No 1 : Adding attribute to entity, or adding new value to a known entity.
\item Type No 2 : Changing or misspecifying the relation between two entities, or interchanging and swapping their roles w.r.t the same relation.
\item Type No 3 : Adding new entities in place of an existing entity, or even otherwise, and mentioning any information about them leaving aside one that purely expresses a no-information stance
\item Type No 4 : Mistakenly claiming knowledge or committing to action about something that the model doesnt really know or cannot act upon
\item Type No 5 : A genuine error in  the logic and inference beyond just new entities, misattributed or swapped roles and relations.
\item Type No 6:  Definitely none of the above, it is something else
\end{itemize}

\section{AMT Annotation Guidelines, Setup and Template}
This section describes the AMT annotation guidelines for OpenDialKG-Eval.

\begin{figure*}[t]
    \centering
    \includegraphics[width=0.85\textwidth]{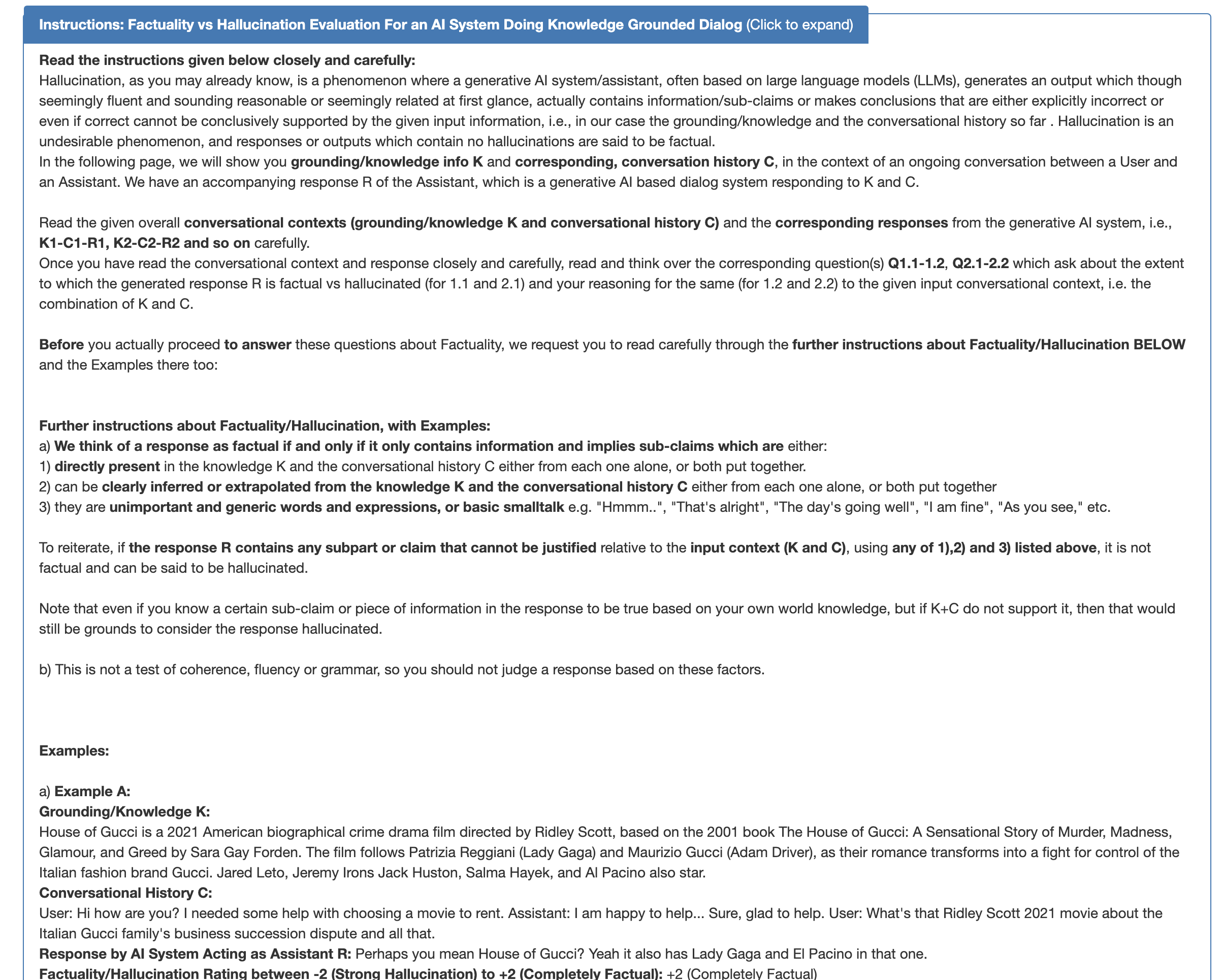}
    \caption{\footnotesize A snapshot of how the initial instructions and examples section of the template would appear to an annotator doing a HIT for our annotation task. } \label{fig:templateSnap}
\vspace{-3mm}
\end{figure*}

A snapshot of the template instructions as seen for an actual example can be viewed in Figure \ref{fig:templateSnap}. Furthermore, we enclose the complete annotation template [including rules and illustrative examples in its contents] in the form of a single .html file included in the Supplementary Materials along with this submission.

Annotators were restricted to be from Anglophone countries (USA, UK, Australia and New Zealand) to ensure a good likelihood of them being native speakers. Further, annotators were restricted to be from among those with a prior approval rate of atleast 98\%.

Annotators were compensated fairly at a rate of 9.3$\$$ per HIT per hour which is well over the minimum wage of 7.25$\$$ per hour in the U.S.A as per Department of Labour estimates for 2023.

We also provide due warning to the annotators not to even inadvertently share any PII or personal information and this is in no way required for our task. We also assure them that time taken etc [nothing beyond the task pertinent annotation] will be used or shared. The disclaimer we include in the template is "Important Disclaimer: Please avoid sharing any personal details or information including PII or demographics anywhere in this study. We will also not be sharing how much time you took to solve this, or what your individual experience profile was. We will merely be using the judgements made about aspects of generated output in relation to input. No other data implicitly or explicitly collected will be shared."

\section{Statistics of OpenDialKG-Eval}
\label{appendix:data_stats}

OpenDialKG-Eval comprises 180 faithful generations and 132 hallucinations, while the development set contains 39 faithful generations and 51 hallucinations. The inter-annotator agreement, measured by Cohen's, stands at 0.583, indicating a moderate level of agreement according to Fleiss's guidelines \cite{fleiss1981statistical} on interpreting Cohen Kappa magnitude.

For the annotation process, we utilized a scale ranging from -2 to +2, excluding 0. Here, -2 represents strong hallucination, and +2 signifies strong faithfulness. Based on this scale, instances scoring greater than 1 or less than -1 were allocated to the test set, with the remaining instances assigned to the development set.

\section{Prompt for Motivating Zero-Shot Detector Experiment in Intro Table 1}
Prompts are shown in Table~\ref{tab:prompt_motivating_exp}.

\begin{table*}[h]
\caption{Prompt template of zero-shot hallucination detector GPT-3.5-turbo for Table~\ref{tab:outdate_performance}. 
} 
\label{tab:prompt_motivating_exp}
\centering
\small
\begin{tabular}{l}
\toprule
<DocumentGivenToAISystem>: \{Input/Document\}</DocumentGivenToAISystem> \\
<SummaryByAISystem>: \{System Output\}</SummaryByAISystem> \\
Is the output Summary generated by the AI System Faithful to the Document given to it?\\ Or is it Hallucinated? (Answer with +1 for Faithful or -1 for Hallucinated): \\
\bottomrule
\end{tabular}
\end{table*}

\section{Prompt for GPT-4 (Internal) Zeroshot Approach}

Prompts are shown in Table~\ref{tab:prompt_gpt4_internal}

\begin{table*}[h]
\caption{Prompt for GPT-4 (Internal) Zeroshot Approach (The Ternary version with Generic, the binary one omits the part concerned with Generic class)
} 
\label{tab:prompt_gpt4_internal}
\centering
\small
\begin{tabular}{l}
\toprule
<PromptGivenToExtBot>: \{Knowledge\}</PromptGivenToExtBot> \\
<ConvHistoryBetweenUserAndExtBot>: \{System Output\}</ConvHistoryBetweenUserAndExtBot> \\
<ResponseByExtBot>: \{System Output\}</ResponseByExtBot> \\

The Response here can be either Faithful to the Context (Prompt and ConvHistory) OR it  
 ecan be hallucinated/contain hallucinations \\ (says something that is contradictory or not \\entirely or close to likely supported by the context). \\
A third possibility is that it says something really generic and not really having a relevant truth value or sufficient relatibility to context,\\ such as smalltalk, obviously \\ true statements amongst other things.\\

Thus a Response can be Faithful, Hallucinated or Generic w.r.t the Prompt given to it and the ConvHistory. \\

Is the output Response given by the ExtBot Faithful to the Prompt given to it and the ConvHistory between User and ExtBot so far?\\ Or is it Hallucinated? Or is it Generic? \\(Answer with 2 for Faithful, 1 for Generic or  0 for Hallucinated): \\
\bottomrule
\end{tabular}
\end{table*}



\section{Experimental Details}
\label{appendix:finetune}
During finetuning, we use batchsize = 4, apply AdamW gradient descent ~\cite{loshchilov2018decoupled} and tune learning rates from the range of $[$1e-3, 1e-4, 1e-5$]$. We run 5 epochs for OpenDialKG and 1 epoch over the BEGIN dataset.
We evaluate our model, HaluEval-based and FADE-based finetuned models using Dev set, choose the best performing learning rate and report the
performances on test set. In addition, we adopt Low Rank Adaptation~\cite{hu2022lora} with r=16, $\alpha$=32, and target\_modules=$[$"q", "v"$]$ during optimization. Our experiments are base on 

BEGIN-Adv. has 8k unreleased data for training, while there are only 1.2k data in BEGIN dev that we can leverage for fine-tuning. In order to generate similar amount of training data, we generate 3 synthetic responses per category for each example in BEGIN Dev set. We adopt temperature 0.5 to avoid repeat generation. 

Wherever pertinent, we provide mean results over two random runs.

\section{Additional Arguments for Cost Efficiency of our Method}
\label{appendix:addedCostEff}
We re-emphasize some key additional points here regarding relative cost efficiency of our approach:

\begin{enumerate}
    \item Inference Time Efficiency: Our findings, presented in Table 2 and Table 3, demonstrate that a fine-tuned T5-base classifier offers significantly lower latency compared to other baseline methods. This is crucial for applications where high latency is not viable.
    \item An additional pragmatic aspect bolstering our relative cost efficiency angle in the long term is the steadily decreasing per-token costs of API-based colossal LLMs like GPT-4, driven by widening adoption, faster accelerators, better compression/distillation and quantization etc. Specifically for GPT-4, we have in the past year seen three major price decrease events, each by a factor of 2+, on June 13th, November 11th and Jan 25th, causing a price decrease of over 8 fold. In contrast, human annotation costs in dollar terms rise albeit very gradually.
\end{enumerate}

\section{Out of Domain Evaluation} 
\label{appendix:ood}
 We conducted additional experiments to evaluate our model's performance in out-of-distribution scenario. Specifically, we utilized the best-performing T5 models, which were finetuned using our dataset generated from OpenDialKG, and applied the model to the BEGIN test set. And we compared with our internal GPT-4 baseline.

\begin{table}[h]
\caption{Detection Results on Out-of-Distribution Data.
} 
\label{tab:ood}
\centering
\small
\begin{tabular}{l|r}
\toprule
Approach & Macro-F1 \\
\midrule
Random baseline & 0.455 \\
Our approach & 0.518 \\
GPT-4 (Internal)	& 0.543 \\
\bottomrule
\end{tabular}
\vspace{-1mm}
\end{table}

Given that the BEGIN dataset features significantly longer knowledge contexts in natural language text, as opposed to the list of triplets from the Freebase knowledge graph found in OpenDialKG, this setup simulates a substantial out-of-distribution evaluation. To adapt our binary classifiers for use with BEGIN, we collapse the “generic” and “not fully attributable” labels into a single “hallucination” category, resulting in a modified binarized test set. The results are as follows in Table \ref{tab:ood}, which indicates that our approach leads to signifantly better performance in comparison to a naive random baseline, and shows competitive performance with GPT-4 based zero-shot detection in out-of-distribution scenario, emphasizing its generalization ability.



\end{document}